# Densely Connected High Order Residual Network for Single Frame Image Super Resolution

**Yiwen Huang**
Department of Computer Science
Wenhua College, Wuhan, China
nickgray0@gmail.com

**Ming Qin**
Department of Computer Science
Wenhua College, Wuhan, China
noahqm@126.com

## Abstract

*Deep convolutional neural networks (DCNN) have been widely adopted for research on super resolution recently, however previous work focused mainly on stacking as many layers as possible in their model, in this paper, we present a new perspective regarding to image restoration problems that we can construct the neural network model reflecting the physical significance of the image restoration process, that is, embedding the a priori knowledge of image restoration directly into the structure of our neural network model, we employed a symmetric non-linear colorspace, the sigmoidal transfer, to replace traditional transfers such as, sRGB, Rec.709, which are asymmetric non-linear colorspaces, we also propose a "reuse plus patch" method to deal with super resolution of different scaling factors, our proposed methods and model show generally superior performance over previous work even though our model was only roughly trained and could still be underfitting the training set.*

## 1. Introduction

Single image super resolution (SISR) has been one of the classic ill-posed problems of image restoration, it tries to reconstruct a high resolution (HR) image from a given low resolution (LR) image. The relationship between LR and HR could be interpreted as LR being the low-passed version of HR, though the kernel of the low-pass filter is generally unknown, most recent studies assume that the kernel would be a Catmull-Rom bicubic filter since it is seemingly the most widespread image resampling algorithm for most image processing applications.

Recent studies show that deep neural networks (DNN) significantly outperform traditional super resolution methods in terms of both objective quality and subjective quality as DNNs come in great capability of non-linear mapping, approximating any measurable function to any precision given that the neural network contains sufficient amount of neurons (parameters). Previous research on solving the super resolution problem with neural networks has always being going along this path focusing on trying to stack more layers in their model, for the fact that the intuition behind this path seems obvious and sound, the capacity of the neural network, the capability of non-linear mapping grows exponentially as we stack more layers to it, however it has been shown that deep neural networks are extremely difficult to train since gradient vanishing has always been haunting the back propagation of deep neural networks. Certain new activation functions, the ReLu family activation functions have been introduced to ameliorate the gradient vanishing problem but the depth of trainable deep neural networks is still limited. This limit has been apparently eliminated with the introduction of ResNet [3] by He *et al.*, although if ResNets are genuine deep neural networks or ensembles of shallower networks [18] remains debatable. Nevertheless, the essential idea that ResNet demonstrates, residual learning through skip connections, makes the construction of neural networks of arbitrary number of layers possible. The big picture that experimenting with different kinds of skip connections and managing to stuff more layers in the neural network model and expecting some performance gain, more or less, seems to give a definite answer to methods of solving super resolution problem with DNNs.

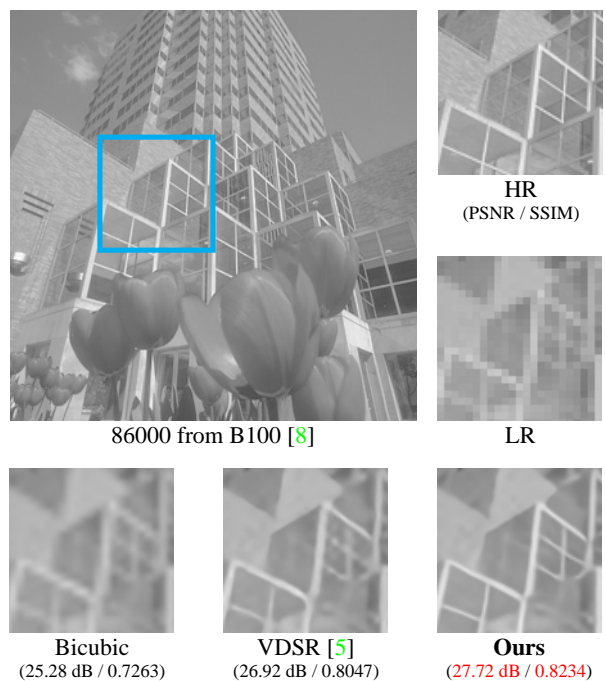

Figure 1: Comparison of our proposed methods with previous work for 4× single image super resolution.

We believe that there should be more, stacking layers cannot be the final conclusion we have toward DNNs for super resolution, the super resolution problem differentiates itself from some other common ill-posed image restoration problems like denoise problem where all components of the image are equally corrupted if the image is polluted by white noise, but different components of the image are not equally corrupted for super resolution. The nature of resampling filters being low-pass filters gives us the rough conclusion that resampling filters tend to retain low frequency components of the image while damaging the high frequency components. The corruption introduced by downsampling could be quantified by information entropy, the aforementioned nature of resampling filters could then be rephrased as downsampling introduces more entropy to high frequency components of the image than to the low frequency components. This assumption has far by now been proven by the observation that the difficulty of reconstructing different components of the image varies greatly, naïve resampling algorithms like bicubic or Lanczos are sophisticated enough to deal with low frequency components, they however produce several artifacts including aliasing, ringing, and probably other more near the bare edges in the reconstructed image, which are high frequency components, indicating these algorithms are not capable of properly reconstructing bare edges. Later work such as sparse representation [1] based methods or early attempts of applying CNNs on super resolution like SRCNN [2], seem to properly reconstruct bare edges in the reconstructed image yet fails to reconstruct finer details and textures in the image. More recent work including VDSR [5], DRRN [13], EDSR [14] *etc*. could restore finer details and structures to some extent though yet still far from an ideal reconstruction. The fact that downsampling introduces more entropy to fine details than to bare edges, more entropy to bare edges than to low frequency flat areas in the LR image accurately explains how reconstructing different components of the image requires models of various complexities.

Global residual learning, which has been widely adopted since VDSR, separates low frequency components from the rest of the image, the network merely has to deal with the lost high frequency components and low frequency components directly pass through the network. This design could be extended to higher order residuals to separate different components of the image more precisely since the term "high frequency components" is relative, bare edges and fine details are both components of higher frequencies than flat areas, yet fine details are components of even higher frequencies than bare edges, which explains why downsampling introduces more entropy to fine details, thus it is more difficult to reconstruct fine details than to reconstruct bare edges. The model we built for this paper and the NTIRE 2018 Super-Resolution challenge [22] contains residual units up to the 5th order.

## 2. Related Works

In this section, we list out 3 models of previous work that are closely related to ours, RED30 [7], DRRN [13] and the winner of the NTIRE 2017 Super-Resolution Challenge [22], EDSR [14]. While RED30 and DRRN seem to resemble our model from the first glance, the principles that RED30, DRRN and our model were built based on are poles apart.

### 2.1. RED30

RED30 [7] is the model of previous work that most closely resembles our model in general, it features high order residuals, an encoder-decoder substructure, both of which our model also contains, but designed based on different principles. The skip connections bridge between each corresponding pair of encoder and decoder in RED30, a higher order residual unit is placed between the encoder block and the decoder block of the current residual unit. Mao *et al.* explained in their paper [7] that this design helps deconvolutional layers, which are the decoders, better reconstruct the image since feature maps generated by the encoder of lower residual units carry more image detail, since they are less filtered, whereas in our model, skip connections are designed based on the entropy that downsampling introduced to each components of the image. Each residual unit in our model also contains an encoder and a decoder, but a higher order residual unit is placed after the decoder, that is, the mapping from LR to HR for a certain component of the image is already complete after the decoder block of a certain residual unit, the higher order residual unit placed after the decoder block takes what is left by the current residual unit which are image components that the current residual unit fails to map to HR, which are components with more entropy than the current residual unit could handle. These components are separated from components with less entropy that the current residual unit could handle and are passed to a more complex higher order residual unit. Mao *et al.* also state in their paper [7] that the skip connections they designed help with back propagation, that is true but we doubt that if such skip connections would actually work well since the gradient still has to go through the deconvolutional layers of all lower order residual units before it reaches the highest order residual unit, which is placed at the exact center of their model.

## 2.2. DRRN

DRRN [13] is also similar to our model in ways that both DRRN and our model employ a recursive substructure. Tai *et al.* [13] introduced global residual learning (GRL) and local residual learning (LRL) in their model, which corresponds to the first and second order residual unit in our model. LRL employs a recursive substructure to reconstruct the second order residual, which is mainly inspired by DRCN. Layers in the recursive substructure share the same weights for each recursion, whereas in our model, the residual unit, which is also applied recursively for the global structure of our model, does not reconstruct a constant order residual recursively. Each recursion in our model generates a unit of higher order residual for components of the image with more entropy than the current residual unit could handle. The residual unit, which is the recursive substructure in our model, does not share weights with another residual unit, since each residual unit aims to reconstruct different components of the image, they do not share features one another, common features shared by components with lower entropy and components with higher entropy, if any, would be separated from features that are exclusive to components with higher entropy by lower order residual units that aim to deal with components with lower entropy.

## 2.3. EDSR

The structure of EDSR [14] does not share any apparent similarities with our model, the essential part of EDSR has a series of ResBlocks wrapped in a global skip connection, it also puts most of its stress on reconstructing the second order residual similar to DRRN [13], except the second order residual is not reconstructed recursively in EDSR, it is repeatedly reconstructed by a series of ResBlocks that do not share any weights with each other. We cast doubt on such design as it is unclear what each ResBlock actually does here, the second order residual reconstruction seems to be divided into all of the ResBlocks wrapped in the global skip connection, it is unclear that if we randomly remove one or more ResBlocks from EDSR, how the super resolution reconstruction results will be affected, whereas at least we have the rough idea of how our reconstruction results will be affected if we remove one or more residual units from our model, we could assert that the reconstruction of very fine details and structures in the image will be affected if we remove residual units of the highest few orders from our model, one major fact that EDSR did inspire us was that Lim *et al.* [14] found batch normalization (BN) layers were unnecessary for super resolution, we do not have any BN layers in our model.

---

[1] http://www.imagemagick.org/Usage/resize/#resize_sigmoidal

## 3. Proposed Methods

In this section, we present the detailed analysis of our work. We introduce sigmoid colorspace conversion and non-ringing Gaussian-Spline resampling algorithm for data preprocessing. We divide our model design into 3 subsections, the residual unit, the global model design and the integration of dense connections [17].

## 3.1. Preprocessing

Recent work on super resolution reconstruction with neural network models has adopted transposed convolutional layers (some may also use the term "deconvolutional layers") or subpixel convolutional layers [9] which enable the model to directly take the downsampled image as the input and output the reconstructed result in target dimensions, these layers seem to render most of the data preprocessing useless, a resampling algorithm is no longer required to first interpolate the downsampled image to obtain the target dimensions before feeding it to the neural network model, however we do not employ these layers in our model for 2 major concerns. First, it has been reported that deconvolutional layers introduce checkerboard artifacts, second, deconvolutional layers and subpixel convolutional layers simply could not be made compatible with the principles that our model is built based on, our proposed model contains multiple residual units to gradually map the corrupted components of the LR image to the corresponding HR components, it does not make sense with these layers which alters the dimensions of the feature maps.

Thus, we have to interpolate the LR image before feeding it to our model, we do not simply blowup the LR image with bicubic as some previous work did for the fact that traditional resampling algorithms are prone to artifacts. We propose 2 effective methods against resampling artifacts that have been introduced for years but never seemed to be formally included in any paper.

**Sigmoidal transfer**[1]: Resampling in non-linear colorspaces, most commonly, in gamma compressed colorspaces (gamma-ignorant resampling) is mathematically incorrect, however it does help with preventing clipping and ringing artifacts while resampling in practical use. All known transfer functions other than the proposed sigmoidal transfer are exponent-like functions which are asymmetric, these functions do not provide any protection against clipping and ringing artifacts on brighter parts of the image whereas the symmetric sigmoidal transfer provides equal protection against clipping and ringing artifacts on both darker and brighter parts of the image, concretely, it is defined as:

$$Y_{linear} = \frac{\frac{1}{1+e^{\beta(\alpha - Y_{sigmoidal})}} - \frac{1}{1+e^{\beta\alpha}}}{\frac{1}{1+e^{\beta(\alpha-1)}} - \frac{1}{1+e^{\beta\alpha}}} \quad (1)$$

α is the inflection point of the sigmoidal transfer, it is equal to 0.5 for the work of this paper and also for any general use, β controls the slope of the sigmoidal transfer, larger value gives a steeper curve. We trained our model on the DIV2K [19] dataset provided by NTIRE 2018, we converted images in the dataset to the linear colorspace before further converted them to the sigmoidal colorspace (β = 8.5) assuming the original transfer was sRGB since the transfer characteristics of the dataset were never released, sRGB transfer seemed to be the most possible transfer out of all alternatives.

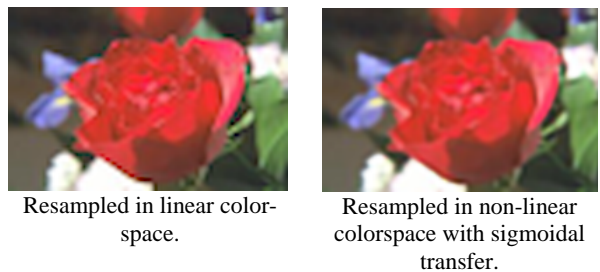

| Resampled in linear colorspace. | Resampled in non-linear colorspace with sigmoidal transfer. |

Figure 2: Demonstration of how sigmoidal transfer helps with preventing certain resampling artifacts.

**Gaussian-Spline resampling kernel**[2]: We adopt spline as the main resampling kernel to obtain the target size for the LR image, the spline kernel, like any kernel with negative slopes, produces ringing artifacts, we pair it with a Gaussian kernel which does not have any negative slope to minimize the ringing artifacts, concretely, for any sample interpolated by the spline kernel, the value of the sample will be clamped to fall between the minimum and maximum of the corresponding 3 × 3 neighborhood generated by the Gaussian kernel.

### 3.2. Residual unit

The residual unit substructure is designed to map certain components of the LR image to the corresponding components of the HR image. Based on the assumption that a certain image could be resolved into a certain combination of the common features that are universally shared for any image, the content of the image should be stored as the coefficients of the feature combination, which should remain untouched when downsampled. The LR and HR image would share the same feature combination coefficients except that the LR image would be a combination of corrupted LR features while the HR image would be the same combination of native HR features, the mapping from LR image to HR image could then be reinterpreted as the mapping from LR features to HR features.

To extract LR features from LR images, we first place an encoder which consists of a few 3 × 3 convolutional layers and activation layers, in the residual unit for such intention, then followed by a symmetric decoder block that maps the extracted LR features to the corresponding HR features, the entire decoder block is wrapped in a skip connection that it merely has to map the difference between LR and HR features. A similar higher order residual unit would be placed after the decoder block if the current residual unit is not the unit of the highest order residual of the model. Then comes the dimensionality reduction block placed at the end of the residual unit to adjust the number of the feature maps of the current residual unit before adding them to the previous residual unit, it holds a 1x1 convolutional layer and an activation layer, the number of filters in the 1x1 convolutional layer is defined by the number of feature maps of the previous residual unit. The dimensionality reduction block could also be used simply for some extra non-linearity if the previous residual unit shares the same number of feature maps with the current residual unit.

---

[2] https://forum.doom9.org/showthread.php?t=145358

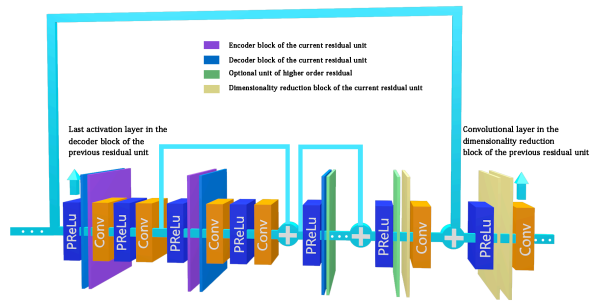

Figure 3: Architecture of one residual unit (*FnL2*) of arbitrary residual order.

The hyperparameters for a residual unit are defined by the number of filters in the convolutional layers of the encoder and decoder block denoted by *Fn* and the number of convolutional and activation layers in the encoder and decoder block denoted by *Ln*. Thus, a residual unit that contains 4 convolutional layers of 256 filters for its encoder and decoder block could be denoted by *F256L4*.

### 3.3. Global model design

The global model design should explicitly reflect the nature and the physical significance that we discussed in Sec. 1 that the global structure of our model should contain substructures that each handles components of the image with certain entropy. We apply the residual unit in our model recursively to construct a high order residual network that each residual unit aims to reconstruct components with more entropy than the previous lower order residual units could properly reconstruct. Any component with more entropy than the current residual unit could properly reconstruct would be separated from components with less entropy that the current residual unit could reconstruct and passed down to residual units of higher orders. Each residual unit should be more or at least equally as complex than previous residual units of lower orders since a higher order residual unit aims to reconstruct components that lower order residual units fail to reconstruct, that is, the components that a higher order residual unit aims to reconstruct contain more entropy than components that lower order residual units aim to reconstruct. The higher order residual unit should thus have a larger capacity for the more intricate mapping from high entropy LR components to the corresponding HR components, concretely, we place 5 residual units in our model, the specifications for each residual unit are inspired by VGG19 [16] net and presented in Table 1.

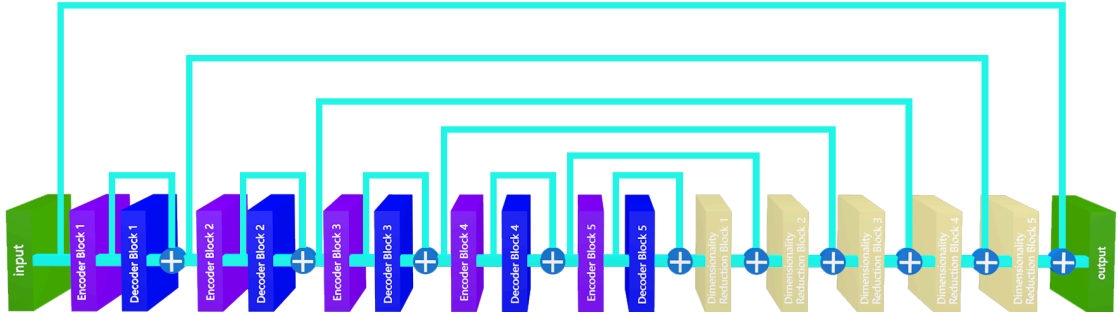

Figure 4: Global architecture of our proposed model without dense connections, building blocks of the *n*th order residual unit are denoted by Encoder Block *n*, Decoder Block *n* and Dimensionality Reduction Block *n*.

| Residual Order | F | L |
|---|---|---|
| 1 | 64 | 2 |
| 2 | 128 | 2 |
| 3 | 256 | 4 |
| 4 | 512 | 4 |
| 5 | 512 | 4 |

Table 1: Specifications for each residual unit.

The first order residual unit is however slightly different from other residual units, the dimensionality reduction block of the first order residual unit does not contain an activation layer, the only convolutional layer in this block also contains only weight parameters, no bias.

### 3.4. Integration of dense connections

We conclude from the current structure of our model that the input of each residual unit gets sparser as the residual order of the unit gets higher since a higher order residual unit takes the residual components that all previous lower order residual units fail to reconstruct yet each lower order residual unit will properly reconstruct some components, which tells us that the residual components passed down to the high order residual unit get sparser every time they go through a residual unit. To avoid the catastrophic contingency that the residual components might get overly sparse and do not carry sufficient amount of valid information for the higher order residual unit to perform a valid reconstruction, we integrate dense connections [17] in our model which directly bridge intermediate feature maps with ample information in lower order residual units to the input of the building blocks of the higher order residual unit, concretely, 2 threads of dense connections are integrated in our model. Let *En* denote the feature maps generated by the encoder block of the *n*th order residual unit, *Dn* denote the feature maps generated by the decoder block of the *n*th order residual unit, specifically, *E0* denote the input LR image itself since the LR image could be interpreted as an untouched set of LR features, the input of the encoder block of the *n*th order residual unit is defined as:

$$\begin{cases} Concatenate(Di), i = 1, 2, ..., n\text{-}1 & \text{if } n>1 \\ E0 & \text{if } n=1 \end{cases} \quad (2)$$

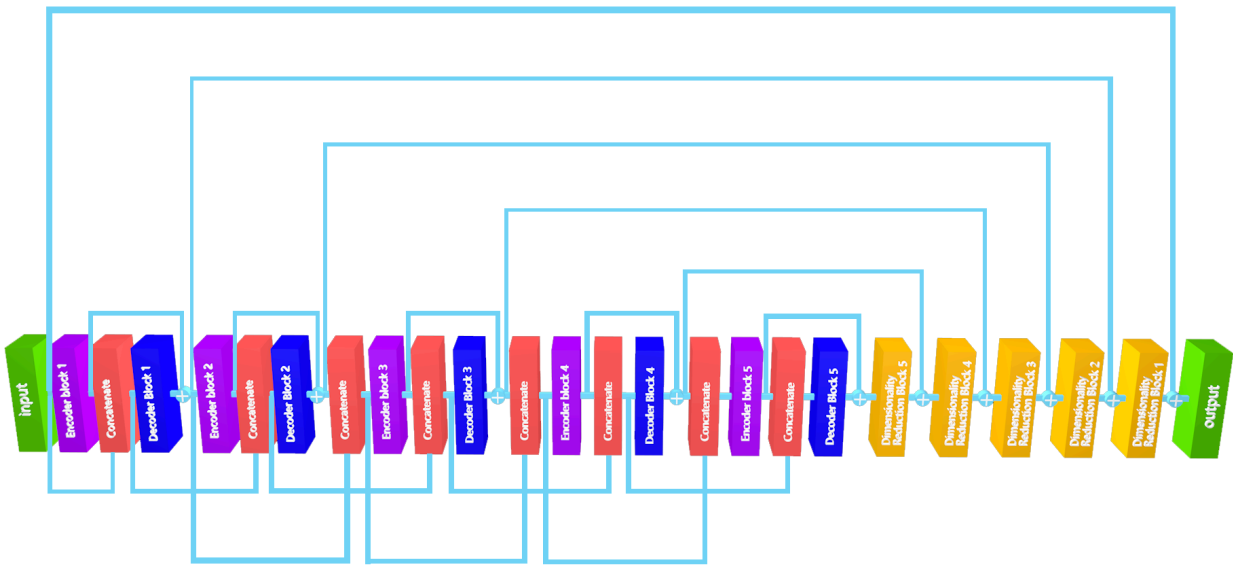

Figure 5: Complete global architecture of our proposed model with the integration of dense connections.

The input of the decoder block of the *n*th order residual unit is defined as:

$$Concatenate(Ei), i = 0, 1, ..., n \quad (3)$$

The encoder block of the residual unit of a certain order takes all available HR features from previous lower order residual units to better extract LR features of the residual components passed down from them, the decoder block takes all available LR features to better map newly extracted LR features of the current residual unit to the corresponding HR features.

## 4. Experiments

### 4.1. Datasets

We train our model on DIV2K [19], a relatively new high-quality dataset containing 800 training images, 100 images for validation and another 100 for testing, provided by the NTIRE workshop. We compare the performance of our model on 4 standard benchmark datasets: Set5 [12], Set14 [11], B100 [8] and Urban100 [4].

### 4.2. Training Details

We optimize our model for logcosh loss, it is more robust than L2 against outliers, the gradient of logcosh loss decays as the loss converges similar to L2, which is generally preferred over loss functions with non-decaying gradient like L1 that might cause severe oscillations as the loss converges unless carefully tweaking the learning rate throughout the training stage.

We perform data augmentation on our entire training set that each pair of LR and HR image are rotated by 90°, 180°, 270° and transposed over the major and counter diagonal, we train our model with grayscale patches of size 159 × 159 from upsampled LR images and the corresponding HR patches, all training samples are normalized to have a zero mean value. We optimize our model with ADAM [6] optimizer setting $\beta_1 = 0.9$, $\beta_2 = 0.999$, $\epsilon = 10^{-8}$. We use mini-batch size of 6 since our GPU memory is limited, a constant learning rate of $2 \times 10^{-4}$.

We implemented our model with Keras 2 and TensorFlow backend. We trained our model on a rented server with 2 NVIDIA GeForce GTX 1080 Ti GPUs for roughly a week since our budget was limited, each model (2× main model and 4× patch model) was only trained for 2 epochs as the models were fairly complex and we did not have more GPUs to speed up the training, our models were highly likely suffering from underfitting for the lack of training. The source code of our model is publicly available online for anyone interested in our work.[3]

### 4.3. Reuse plus patch for larger scaling factors

Inspired by LapSRN [15], we divide super resolution reconstruction of larger scaling factors into multiple stages of 2× super resolution reconstruction, concretely for the NTIRE 2018 Super-Resolution Challenge [22], we participated in track 1 that aims to reconstruct a LR image 8× downsampled from the HR image by Catmull-Rom bicubic, this could be divided into 3 stages of 2× super resolution reconstruction. Since each stage of 2× reconstruction should be quite similar, we trained our main model for 2× reconstruction based on such assumption, a patch model was applied for 2× to 4× reconstruction and a patch model for 4× to 8× likewise, however, the last patch model for 8× reconstruction was only planned but not implemented since our budget was limited. Let *F2* denote our main model of 2× reconstruction, *P4* denote the patch model for 4× reconstruction and *P8* for 8× reconstruction. *P4* would be trained to minimize the difference between

$$F2(F2(I_{LR})) \quad (4)$$

and $I_{HR}$, *P8* would be trained to minimize the difference between

$$P4(F2(P4(F2(F2(I_{LR}))))) \quad (5)$$

and $I_{HR}$, this method could be further extended if necessary, *P16* for instance, if required, would be trained to fit between

$$P8(P4(F2(P8(P4(F2(P4(F2(F2(I_{LR}))))))))) \quad (6)$$

and $I_{HR}$. We used the same neural network structure of 5 residual units for both our 2× main model and 4× patch

---

[3] https://github.com/ccook0/NTIRE2018

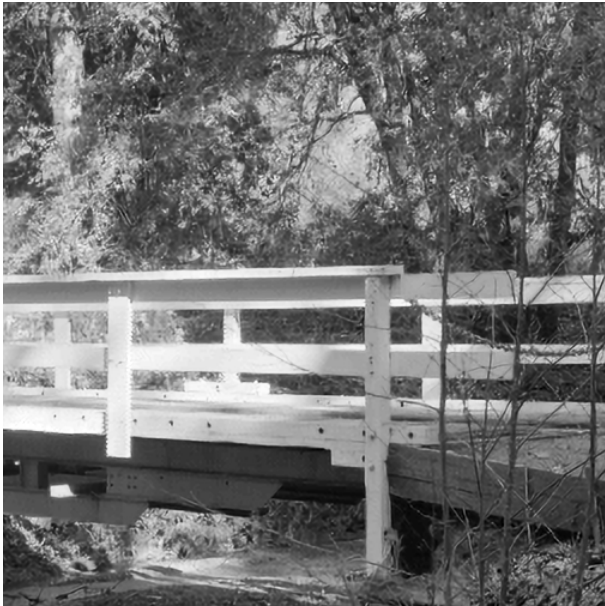
Result of 2× super resolution generated by our overfitted lite model.

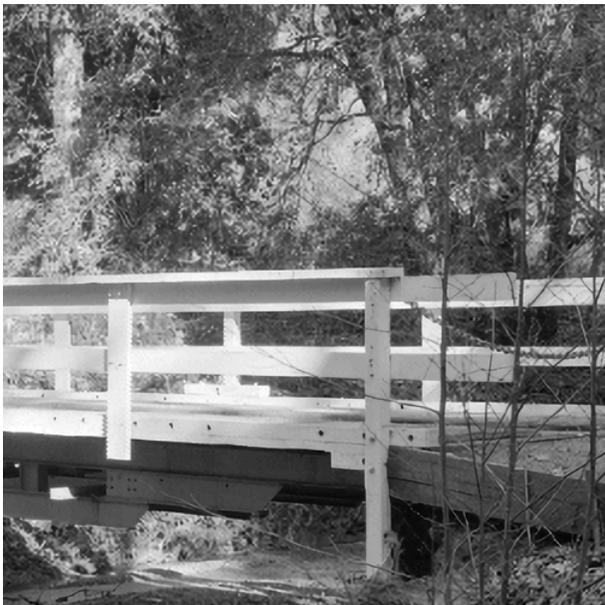
Result of 2× super resolution generated by our proposed final model trained on the entire DIV2K [19] dataset.

Figure 6: Comparison of different behaviors our model exhibits when trained on datasets of different sizes.

model for the NTIRE 2018 Super-Resolution Challenge [22].

### 4.4. Generating photo-realistic reconstructions without GAN or perceptual loss

We discovered by chance that our model exhibited behaviors similar to SRGAN [20] generating photo-realistic reconstructions when overfitted to a small but high quality dataset. We created a lite version of our final model which only had the first 3 residual units of our final model and trained it with merely 4 images in the DIV2K [19] dataset, 0003, 0007, 0015 and 0025 from DIV2K. The mini dataset was fully augmented like how we trained our final model with the complete DIV2K dataset, the training error was near zero (PSNR > 40 dB) that it seemed our lite model just memorized the entire mini dataset. The lite model was simply optimized for logcosh loss like our final model, no perceptual loss [21] or GAN structure was involved. We then tested our overfitted lite model on a few new samples and the performance was predictably poor in terms of PSNR, the generated results, however, were not exactly garbage results in terms of subjective quality. The generated images were quite photo-realistic and rich in details and fine textures, in some ways, similar to the results of SRGAN. It has been suggested in the paper [20] of SRGAN that the essential mechanism of generating photo-realistic images is learning the exact distribution of HR images. It seems our model would learn the distribution of HR images autonomously without the guidance of a discriminator network or perceptual loss given that the training set is about the just right size for the model to perfectly memorize all samples in the training set. Further research and discussions are required to investigate if such phenomenon is unique to our model or universal for all generative models given such conditions.

### 4.5. Benchmark Results

The quantitative evaluation results are listed in Table 2. We compare our model with previous work including A+ [10], SRCNN [2], VDSR [5], RED30 [7], DRRN [13] and EDSR [14]. Our model and methods show generally superior performance over previous work, however, our model does not top EDSR on the Urban100 [4] dataset, we argue that the performance of our model could be further improved with adequate training.

## 5. Conclusion

In this paper, we presented a fresh perspective that instead of blindly stacking layers, neural network models could be designed to reflect the physical significance or the a priori knowledge of specific applications. Concretely for our work, we proposed a neural network model and supporting methods for single image super resolution (SISR) designed based on the information entropy of each components of the image and achieved the state-of-the-art performance even when the model was not thoroughly trained.

We argue that our model design could also be applied on other image restoration tasks with minor modifications as long as different components of the image are not equally corrupted, typically, deblurring and compression artifacts removal.

| Dataset | Scale | A+ [10] | SRCNN [2] | VDSR [5] | RED30 [7] | DRRN [13] | EDSR [14] | **Ours** |
|---|---|---|---|---|---|---|---|---|
| Set5 | ×2 | 36.54 / 0.9544 | 36.66 / 0.9542 | 37.53 / 0.9587 | 37.66 / 0.9599 | 37.74 / 0.9591 | 38.20 / 0.9606 | 39.55 / 0.9665 |
| | ×4 | 30.28 / 0.8603 | 30.48 / 0.8628 | 31.35 / 0.8838 | 31.51 / 0.8869 | 31.68 / 0.8888 | 32.62 / 0.8984 | 33.62 / 0.9032 |
| Set14 | ×2 | 32.28 / 0.9056 | 32.42 / 0.9063 | 33.03 / 0.9124 | 32.94 / 0.9144 | 33.23 / 0.9136 | 34.02 / 0.9204 | 34.65 / 0.9264 |
| | ×4 | 27.32 / 0.7491 | 27.49 / 0.7503 | 28.01 / 0.7674 | 27.86 / 0.7718 | 28.21 / 0.7720 | 28.94 / 0.7901 | 29.72 / 0.8001 |
| B100 | ×2 | 31.21 / 0.8863 | 31.36 / 0.8879 | 31.90 / 0.8960 | 31.99 / 0.8974 | 32.05 / 0.8973 | 32.37 / 0.9018 | 33.24 / 0.9076 |
| | ×4 | 26.82 / 0.7087 | 26.90 / 0.7101 | 27.29 / 0.7251 | 27.40 / 0.7290 | 27.38 / 0.7284 | 27.79 / 0.7437 | 28.63 / 0.7556 |
| Urban100 | ×2 | 29.20 / 0.8938 | 29.50 / 0.8946 | 30.76 / 0.9140 | - / - | 31.23 / 0.9188 | 33.10 / 0.9363 | 32.42 / 0.9272 |
| | ×4 | 24.32 / 0.7183 | 24.52 / 0.7221 | 25.18 / 0.7524 | - / - | 25.44 / 0.7638 | 26.86 / 0.8080 | 26.70 / 0.7823 |

Table 2: Benchmark results for ×2 and ×4 super resolution (PSNR(dB) / SSIM). Red indicates the best performance and blue indicates the second best.

## References

[1] J. Yang, J. Wright, T. S. Huang, and Y. Ma. Image super-resolution via sparse representation. *IEEE Transactions on Image Processing*, 19(11):2861–2873, 2010. 2

[2] C. Dong, C. C. Loy, K. He, and X. Tang. Learning a deep convolutional network for image super-resolution. In *ECCV 2014*. 2, 7, 8

[3] K. He, X. Zhang, S. Ren, and J. Sun. Deep residual learning for image recognition. In *CVPR 2016*. 1

[4] J.-B. Huang, A. Singh, and N. Ahuja. Single image super-resolution from transformed self-exemplars. In *CVPR 2015*. 6, 7

[5] J. Kim, J. Kwon Lee, and K. M. Lee. Accurate image super-resolution using very deep convolutional networks. In *CVPR 2016*. 1, 2, 7, 8

[6] D. Kingma and J. Ba. Adam: A method for stochastic optimization. In *ICLR 2014*. 6

[7] X. Mao, C. Shen, and Y.-B. Yang. Image restoration using very deep convolutional encoder-decoder networks with symmetric skip connections. In *NIPS 2016*. 2, 7, 8

[8] D. Martin, C. Fowlkes, D. Tal, and J. Malik. A database of human segmented natural images and its application to evaluating segmentation algorithms and measuring ecological statistics. In *ICCV 2001*. 1, 4

[9] W. Shi, J. Caballero, F. Huszár, J. Totz, A. P. Aitken, R. Bishop, D. Rueckert, and Z. Wang. Real-time single image and video super-resolution using an efficient sub-pixel convolutional neural network. In *CVPR 2016*. 3

[10] R. Timofte, V. De Smet, and L. Van Gool. A+: Adjusted anchored neighborhood regression for fast super-resolution. In *ACCV 2014*. 7, 8

[11] R. Zeyde, M. Elad, and M. Protter. On single image scale-up using sparse-representations. In *Proceedings of the International Conference on Curves and Surfaces*, 2010. 6

[12] M. Bevilacqua, A. Roumy, C. Guillemot, and M. L. Alberi-Morel. Low-complexity single-image super-resolution based on nonnegative neighbor embedding. In *BMVC 2012*. 6

[13] Y. Tai, J. Yang, and X. Liu. Image super-resolution via deep recursive residual network. In *CVPR 2017*. 2, 3, 7, 8

[14] B. Lim, S. Son, H. Kim, S. Nah, and K. M. Lee. Enhanced deep residual networks for single image super-resolution. In *CVPR 2017 Workshops*. 2, 3, 7, 8

[15] W. S. Lai, J. B. Huang, N. Ahuja, and M. H. Yang. Deep laplacian pyramid networks for fast and accurate super-resolution. In *CVPR 2017*. 6

[16] K. Simonyan and A. Zisserman. Very deep convolutional networks for large-scale image recognition. In *ICLR 2015*. 5

[17] G. Huang, Z. Liu, K. Q. Weinberger, and L. van der Maaten. Densely connected convolutional networks. In *CVPR 2017*. 3, 6

[18] A. Veit, M. Wilber and S. Belongie. Residual networks behave like ensembles of relatively shallow networks. In *NIPS 2016*. 1

[19] E. Agustsson and R. Timofte. NTIRE 2017 challenge on single image super-resolution: dataset and study. In *CVPR 2017 Workshops*. 4, 6, 7

[20] C. Ledig, L. Theis, F. Huszár, J. Caballero, A. Cunningham, A. Acosta, A. Aitken, A. Tejani, J. Totz, Z. Wang, et al. Photo-realistic single image super-resolution using a generative adversarial network. In *CVPR 2017*. 7

[21] J. Johnson, A. Alahi, and L. Fei-Fei. Perceptual losses for real-time style transfer and super-resolution. In *ECCV 2016*. 7

[22] R. Timofte, E. Agustsson, L. Van Gool, M.-H. Yang, L. Zhang, et al. Ntire 2017 challenge on single image super-resolution: Methods and results. In *CVPR 2017 Workshops*. 2, 6, 7